\documentclass[10pt,twocolumn,letterpaper]{article}

\usepackage{cvpr}
\usepackage{times}
\usepackage{epsfig}
\usepackage{graphicx}
\usepackage{amsmath}
\usepackage{amssymb}

\usepackage[pagebackref=true,breaklinks=true,letterpaper=true,colorlinks,bookmarks=false]{hyperref}

\cvprfinalcopy 

\def\cvprPaperID{2559} 

\newcommand*{\mb}[1]{\mathbf{#1}}
\def\be#1\ee{\begin{align}#1\end{align}}
\def\bea#1\eea{\begin{eqnarray}#1\end{eqnarray}}
\def\ba#1\ea{\begin{align*}#1\end{align*}}
\def\bs#1\es{\begin{equation}\begin{split}#1\end{split}\end{equation}}

\ifcvprfinal\fi
\begin{document}

\title{Deep Pyramidal Residual Networks}

\author{Dongyoon Han\thanks{These two authors contributed equally.}\\
EE, KAIST\\
{\tt\small dyhan@kaist.ac.kr}
\and
Jiwhan Kim\footnotemark[1]\\
EE, KAIST\\
{\tt\small jhkim89@kaist.ac.kr}
\and
Junmo Kim\\
EE, KAIST\\
{\tt\small junmo.kim@kaist.ac.kr}
}
\maketitle

\begin{abstract}
 Deep convolutional neural networks (DCNNs) have shown remarkable performance in image classification tasks in recent years. Generally, deep neural network architectures are stacks consisting of a large number of convolutional layers, and they perform downsampling along the spatial dimension via pooling to reduce memory usage. Concurrently, the feature map dimension (i.e., the number of channels) is sharply increased at downsampling locations, which is essential to ensure effective performance because it increases the diversity of high-level attributes. This also applies to residual networks and is very closely related to their performance. In this research, instead of sharply increasing the feature map dimension at units that perform downsampling, we gradually increase the feature map dimension at all units to involve as many locations as possible. This design, which is discussed in depth together with our new insights, has proven to be an effective means of improving generalization ability. Furthermore, we propose a novel residual unit capable of further improving the classification accuracy with our new network architecture. Experiments on benchmark CIFAR-10, CIFAR-100, and ImageNet datasets have shown that our network architecture has superior generalization ability compared to the original residual networks.

 {\small Code is available at https://github.com/jhkim89/PyramidNet}
\end{abstract}


\section{Introduction}
The emergence of deep convolutional neural networks (DCNNs) has greatly contributed to advancements in solving complex tasks~\cite{alexnet, overfeat, decaf, rcnn, FCN} in computer vision with significantly improved performance. Since the proposal of LeNet~\cite{Lenet}, which introduced the use of deep neural network architectures for computer vision tasks, the advanced architecture AlexNet~\cite{alexnet} was selected as the winner of the 2012 ImageNet competition~\cite{ImageNet} by a large margin over traditional methods. Subsequently, ZF-net~\cite{zfnet}, VGG~\cite{VGG}, GoogleNet~\cite{GoogleNet}, Residual Networks~\cite{resnet, preresnet}, and Inception Residual Networks~\cite{InceptionResnet} were successively proposed to demonstrate advances in network architectures. In particular, Residual Networks (ResNets)~\cite{resnet, preresnet} leverage the concept of shortcut connections~\cite{Highway} inside a proposed residual unit for residual learning, to make it possible to train much deeper network architectures. Deeper network architectures are known for their superior performance, and these network architectures commonly have deeply stacked convolutional filters with nonlinearity \cite{VGG,GoogleNet}.

With respect to feature map dimension, the conventional method of stacking several convolutional filters is to increase the dimension while decreasing the size of feature maps by increasing the strides of the filters or poolings. This is the widely adopted method of controlling the size of feature maps, because extracting the diversified high-level attributes with the increased feature map dimension is very effective for classification tasks. Architectures such as those of AlexNet~\cite{alexnet} and VGG~\cite{VGG} utilize this method of increasing the feature map dimension to construct their network architectures. The most successful deep neural network, ResNets~\cite{resnet,preresnet}, which was introduced by He~\textit{et al.}~\cite{resnet}, also follows this approach for filter stacking.

\begin{figure*}[ht]
\small
\centering
\begin{tabular}{c}
\includegraphics[width=170mm, height=45mm]{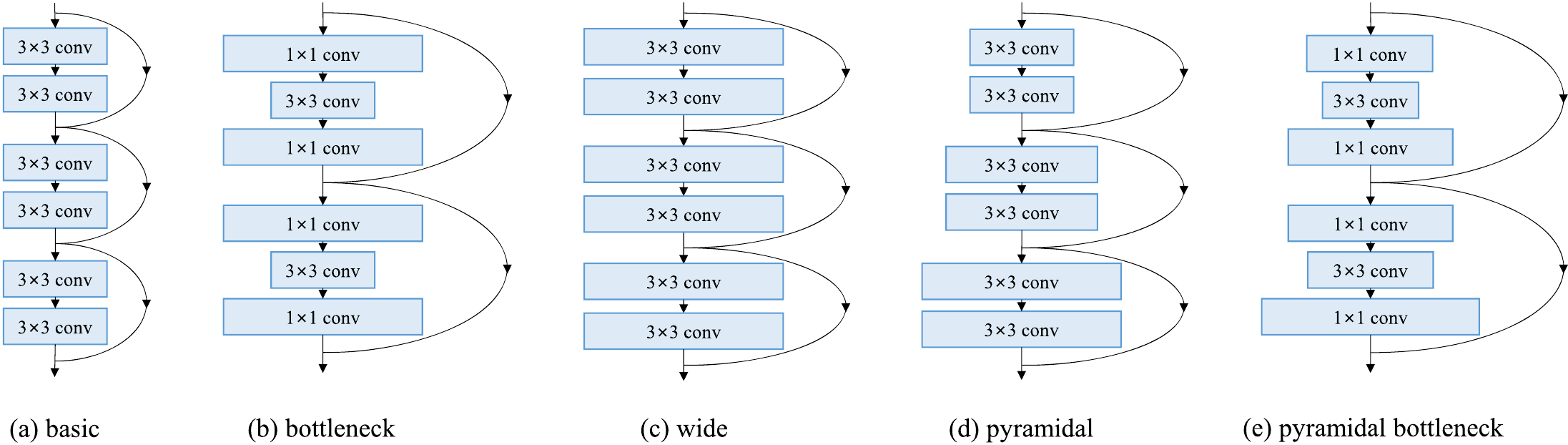}
\end{tabular}
\caption{Schematic illustration of (a) basic residual units~\cite{resnet}, (b) bottleneck residual units~\cite{resnet}, (c) wide residual units~\cite{wideresnet}, (d) our pyramidal residual units, and (e) our pyramidal bottleneck residual units.}
\label{fig:pyramid}
\end{figure*}

According to the research of Veit {\it et al.}~\cite{ensemble}, ResNets are considered to behave as ensembles of relatively shallow networks. These researchers showed that the deletion of an individual residual unit from ResNets, i.e., such that only a shortcut connection remains, does not significantly affect the overall performance, proving that deleting a residual unit is equivalent to deleting some shallow networks in the ensemble networks. Contrary to this, deleting a single layer in plain network architectures such as a VGG-network~\cite{VGG} damages the network by causing additional severe errors.

However, in the case of ResNets, it was found that deleting the building blocks in a residual unit with downsampling, where the feature map dimension is doubled, still increases the classification error by a significant margin. Interestingly, when the residual net is trained using a stochastic depth~\cite{stochasticdepth}, it was found that deleting the blocks with downsampling does not degrade the classification performance, as shown in Figure~8 in~\cite{ensemble}. One may think that this phenomenon is related to the overall improvement in the classification performance enabled by stochastic depth.

Motivated by the ensemble interpretation of residual networks in Veit et al.~\cite{ensemble} and the results with stochastic depth~\cite{stochasticdepth}, we devised another method to handle the phenomenon associated with deleting the downsampling unit. In the proposed method, the feature map dimensions are increased at all layers to distribute the burden concentrated at locations of residual units affected by downsampling, such that it is equally distributed across all units. It was found that using the proposed new network architecture, deleting the units with downsampling does not degrade the performance significantly.
In our paper, we refer to this network architecture as a deep ``pyramidal" network and a ``pyramidal'' residual network with a residual-type network architecture. This reflects the fact that the shape of the network architecture can be compared to that of a pyramid. That is, the number of channels gradually increases as a function of the depth at which the layer occurs, which is similar to a pyramid structure of which the shape gradually widens from the top downwards. This structure is illustrated in comparison to other network architectures in Figure~\ref{fig:pyramid}. The key contributions are summarized as follows:

\begin{itemize}
\item A deep pyramidal residual network (PyramidNet) is introduced. The key idea is to concentrate on the feature map dimension by increasing it gradually instead of by increasing it sharply at each residual unit with downsampling. In addition, our network architecture works as a mixture of both plain and residual networks by using zero-padded identity-mapping shortcut connections when increasing the feature map dimension.
\item A novel residual unit is also proposed, which can further improve the performance of ResNet-based architectures (compared with state-of-the-art network architectures).

\end{itemize}

The remainder of this paper is organized as follows. Section~\ref{sec:architecture} presents our PyramidNets and introduces a novel residual unit that can further improve ResNet. Section~\ref{sec:discussions} closely analyzes our PyramidNets via several discussions. Section~\ref{section:exp} presents experimental results and comparisons with several state-of-the-art deep network architectures. Section~\ref{sec:conclusion} concludes our paper with suggestions for future works.

\section{Network Architecture}
\label{sec:architecture}
In this section, we introduce the network architectures of our PyramidNets. The major difference between PyramidNets and other network architectures is that the dimension of channels gradually increases, instead of maintaining the dimension until a residual unit with downsampling appears. 
A schematic illustration is shown in Figure~\ref{fig:pyramid}~(d) to facilitate understanding of our network architecture.

\subsection{Feature Map Dimension Configuration}
Most deep CNN architectures~\cite{resnet, preresnet, alexnet, VGG, GoogleNet, zfnet} utilize an approach whereby feature map dimensions are increased by a large margin when the size of the feature map decreases, and feature map dimensions are not increased until they encounter a layer with downsampling. In the case of the original ResNet for CIFAR datasets~\cite{cifar}, the number of feature map dimensions $D_k$ of the $k$-th residual unit that belongs to the $n$-th group can be described as follows:
\be
     D_k    =  \begin{cases}
                    16, & \quad \text{if $n(k) = 1$},\\
                    16\cdot2^{n(k)-2}, & \quad \text{if $n(k) \geq 2$},
                \end{cases}
\ee
in which $n(k)\in\{1,2,3,4\}$ denotes the index of the group to which the k-th residual unit belongs. The residual units that belong to the same group have an equal feature map size, and the $n$-th group contains $N_n$ residual units. In the first group, there is only one convolutional layer that converts an RGB image into multiple feature maps. For the $n$-th group, after $N_n$ residual units have passed, the feature size is downsampled by half and the number of dimensions is doubled.
\begin{figure}[t]
\small
\begin{center}
\begin{tabular}{ccc}
\includegraphics[width = 0.11\textwidth, height = 30mm]{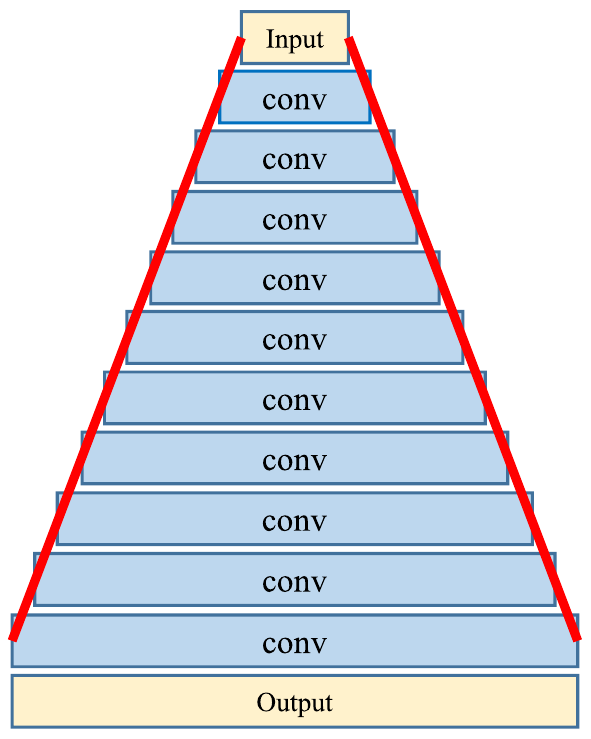} &
\includegraphics[width = 0.155\textwidth, height = 30mm]{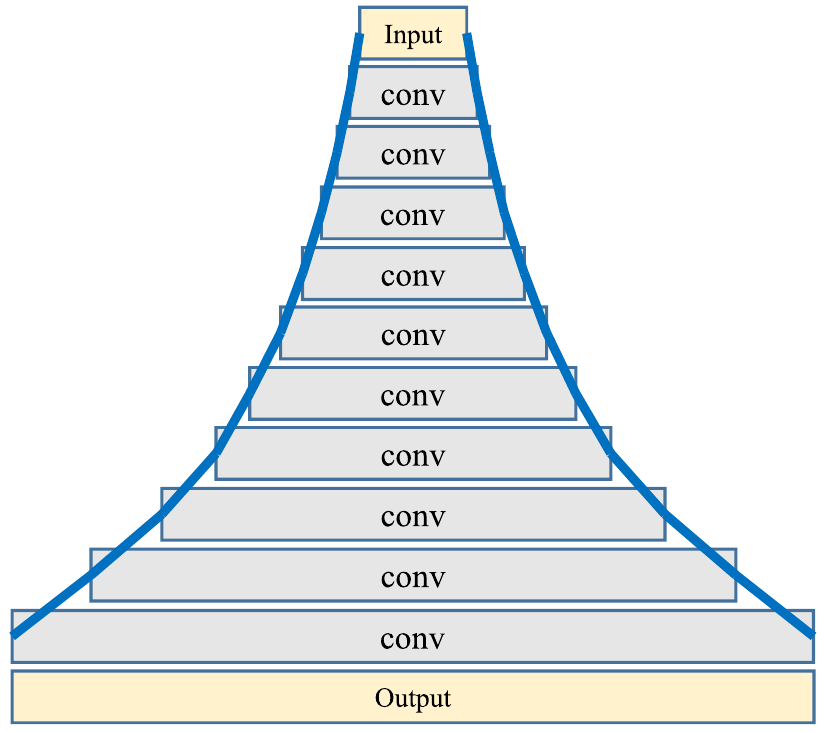} &
\includegraphics[width = 0.155\textwidth, height = 30mm]{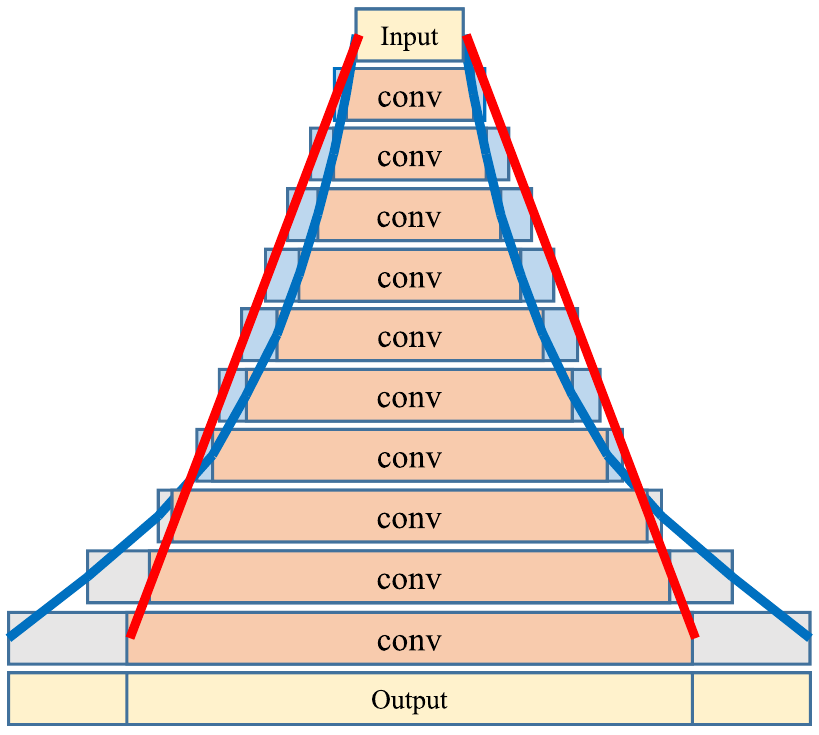} \\
(a) & (b) & (c)
\end{tabular}
\end{center}
\caption{Visual illustrations of (a) additive PyramidNet, (b) multiplicative PyramidNet, and (c) a comparison of (a) and (b).}
\label{fig:addmul}
\end{figure}
We propose a method of increasing the feature map dimension as follows:
\bs
D_{k} =  \begin{cases}
                    16, &  \text{if $k = 1$},\\
                    \lfloor D_{k-1}+  \alpha /N \rfloor, &  \text{if $2 \leq k \leq N+1$},
                \end{cases}
                \label{eq:dimension}
\es
in which $N$ denotes the total number of residual units, defined as $N=\sum_{n=2}^4 N_n$. The dimension is increased by a step factor of $\alpha/N$, and the output dimension of the final unit of each group becomes $16 + (n-1)\alpha/3$ with same number of residual units in each group. The details of our network architecture are presented in Table~\ref{table:structure}.

The above equations are based on an addition-based widening step factor $\alpha$ for increasing dimensions. However, of course, multiplication-based widening (i.e., the process of multiplying by a factor to increase the channel dimension geometrically) presents another possibility for creating a pyramid-like structure. Then, eq.\eqref{eq:dimension} can be transformed as follows:
\bs
        D_{k} =  \begin{cases}
                    16, &  \text{if $k = 1$},\\
                    \lfloor D_{k-1}\cdot \alpha^{\frac{1}{N}} \rfloor,  &  \text{if $2 \leq k \leq N+1$}.
                \end{cases}
                \label{eq:dimension2}
\es
The main difference between additive and multiplicative PyramidNets is that the feature map dimension of an additive network gradually increases linearly, whereas the dimension of a multiplicative network increases geometrically. That is, the dimension slowly increases in input-side layers and sharply increases in output-side layers. This process is similar to that of the original deep network architectures such as VGG~\cite{VGG} and ResNet~\cite{resnet}. The visual illustrations of additive and multiplicative PyramidNets are shown in Figure~\ref{fig:addmul}. In this paper, we compare the performance of both of these dimension-increasing approaches by comparing an additive PyramidNet (eq.~\eqref{eq:dimension}) and a multiplicative PyramidNet (eq.~\eqref{eq:dimension2}) in section~\ref{section:exp}.

\begin{table}[t]
\footnotesize
\begin{center}
\begin{tabular}{|c|c|c|}
\hline
Group & Output size & Building Block\\
\hline\hline
conv 1 & 32$\times$32 & $\left[ \textrm{3}\times\textrm{3}, \textrm{16} \right]$\\
\hline
conv 2 & 32$\times$32 & $\left[ \begin{array}{c} \textrm{3}\times\textrm{3},  \lfloor \textrm{16} + \alpha (k-1)/N \rfloor \\ \textrm{3}\times\textrm{3},  \lfloor \textrm{16} + \alpha (k-1)/N \rfloor \end{array}\right]\times N_2$\\
\hline
conv 3 & 16$\times$16 & $\left[ \begin{array}{c} \textrm{3}\times\textrm{3}, \lfloor \textrm{16} + \alpha (k-1)/N \rfloor\\ \textrm{3}\times\textrm{3},  \lfloor \textrm{16} + \alpha (k-1)/N \rfloor \end{array}\right]\times N_3$\\
\hline
conv 4 & 8$\times$8 & $\left[ \begin{array}{c} \textrm{3}\times\textrm{3}, \lfloor \textrm{16} + \alpha (k-1)/N \rfloor\\ \textrm{3}\times\textrm{3}, \lfloor  \textrm{16} + \alpha (k-1)/N \rfloor \end{array}\right]\times N_4$\\
\hline
avg pool & 1$\times$1 & $\left[8\times8, 16+\alpha\right]$\\
\hline
\end{tabular}
\end{center}
\caption{Structure of our PyramidNet for benchmarking with CIFAR-10 and CIFAR-100 datasets. $\alpha$ denotes the widening factor, and $N_n$ signifies the number of blocks in a group. Downsampling is performed at conv3\_1 and conv4\_1 with a stride of 2. 
}
\label{table:structure}
\end{table}
\subsection{Building Block}
The building block (i.e., the convolutional filter stacks with ReLUs and BN layers) in a residual unit is the core of ResNet-based architectures. It is obvious that in order to maximize the capability of the network architecture, designing a good building block is essential. As shown in Figure~\ref{fig:resnets}, the layers can be stacked in various manners to construct a single building block. We found the building block shown in Figure~\ref{fig:resnets}~(d) to be the most promising, and therefore we included this structure as building block in our PyramidNets. The discussion of this matter is continued in the following section.

In terms of shortcut connections, many researchers either use those based on identity mapping, or those employing convolution-based projection. However, as the feature map dimension of PyramidNet is increased at every unit, we can only consider two options: zero-padded identity-mapping shortcuts, and projection shortcuts conducted by 1$\times$1 convolutions. However, as mentioned in the work of He {\it et al.}~\cite{preresnet}, the 1$\times$1 convolutional shortcut produces a poor result when there are too many residual units, i.e., this shortcut is unsuitable for very deep network architectures. Therefore, we select zero-padded identity-mapping shortcuts for all residual units. Further discussions about the zero-padded shortcut are provided in the following section.

\section{Discussions}
\label{sec:discussions}
In this section, we present an in-depth study of the architecture of our PyramidNet, together with the proposed novel residual units. The experiments we include here support the study and confirm that insights obtained from our network architecture can further improve the performance of existing ResNet-based architectures.

\subsection{Effect of PyramidNet}
According to the work of Veit {\it et al.}~\cite{ensemble}, ResNets can be viewed as ensembles of relatively shallow networks, supported by the observation that deleting an individual building block in a residual unit of ResNets incurs minor classification loss, whereas removing layers from plain networks such as VGG~\cite{VGG} severely reduces the classification rate. However, in both original and pre-activation ResNets~\cite{resnet,preresnet}, another noteworthy aspect is that deleting the units with downsampling (and doubling the feature dimension) still degrades performance by a large margin~\cite{ensemble}. Meanwhile, when a stochastic depth~\cite{stochasticdepth} is applied, this phenomenon is not observed, and the performance is also improved, according to the experiment of Veit~{\it et al.}~\cite{ensemble}. The objective of our PyramidNet is to resolve this phenomenon differently, by attempting to gradually increase the feature map dimension instead of doubling it at one of the residual units and to evenly distribute the burden of increasing the feature maps. We observed that our PyramidNet indeed resolves this phenomenon and at the same time improves overall performance. We further analyze the effect of our PyramidNet by comparing it against the pre-activation ResNet, with the following experimental results.
\begin{figure}[t]
\begin{center}
\begin{tabular}{c}
\includegraphics[trim = 37mm 83mm 31mm 87mm, clip, width=0.42\textwidth, height=0.3\textwidth]{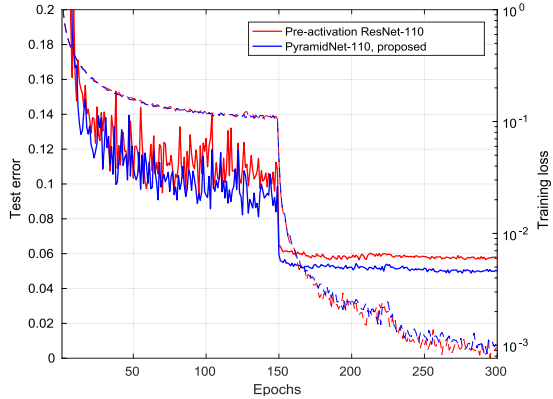}
\end{tabular}
\end{center}
\caption{Performance comparison between the pre-activation ResNet~\cite{preresnet} and our PyramidNet, using CIFAR datasets. Dashed and solid lines denote the training loss and test error, respectively.}
\label{fig:curves}
\end{figure}
First, we compare the training and test error curves of our PyramidNet with those of the pre-activation ResNet~\cite{preresnet} in Figure~\ref{fig:curves}. The standard pre-activation ResNet with 110 layers is used for comparison. For our PyramidNet, we used a depth of 110 layers with a widening factor of $\alpha=48$; it had the same number of parameters (1.7M) as the pre-activation ResNet to allow for a fair comparison. The results indicate that our PyramidNet has superior test accuracy, thereby confirming its greater ability to generalize compared to existing deep networks.

\begin{figure}[t]
\begin{center}
\begin{tabular}{cc}
\includegraphics[trim = 37mm 94mm 44mm 100mm, clip,  width=0.225\textwidth, height=0.18\textwidth]{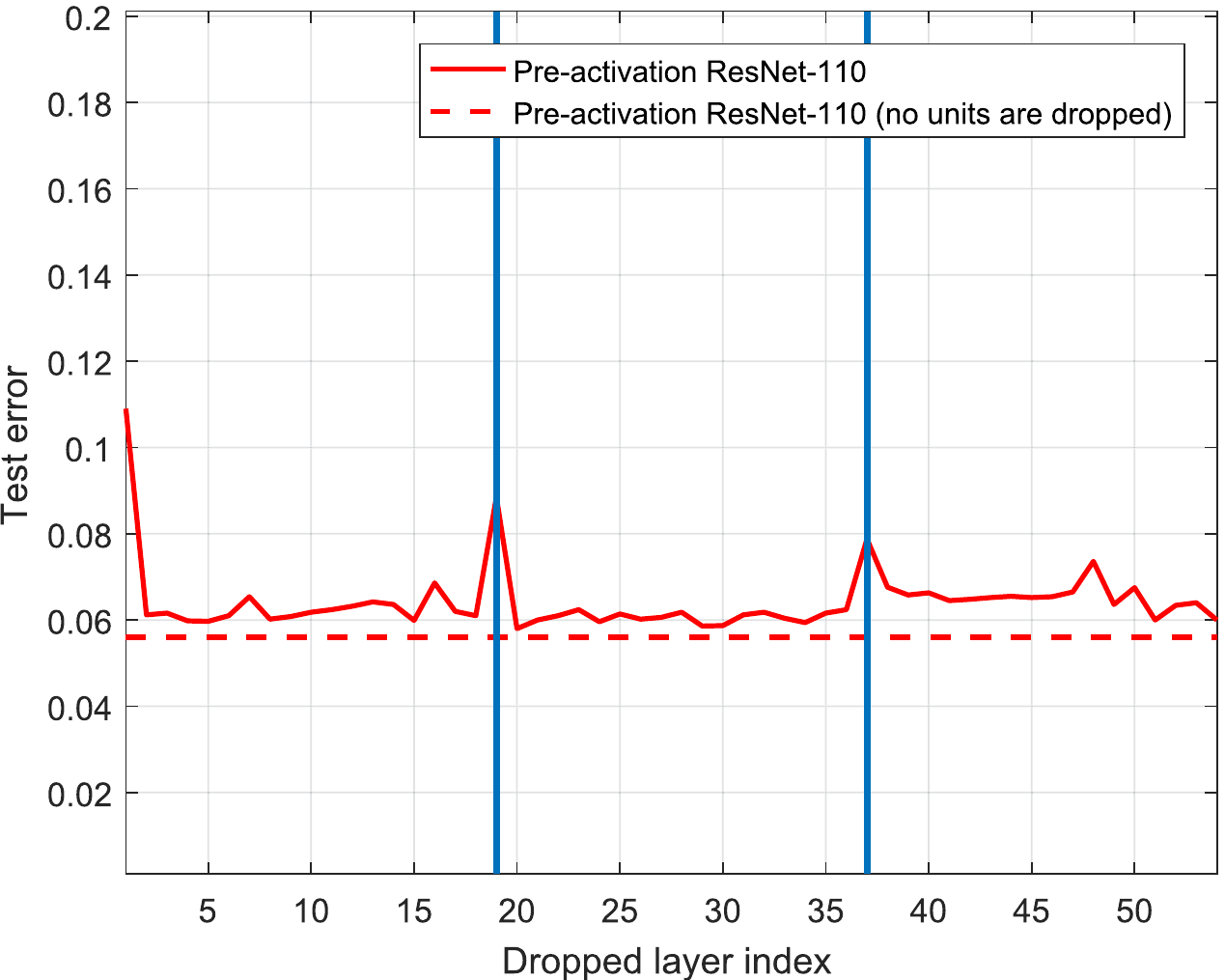} \
\includegraphics[trim = 37mm 94mm 44mm 100mm, clip,  width=0.225\textwidth, height=0.18\textwidth]{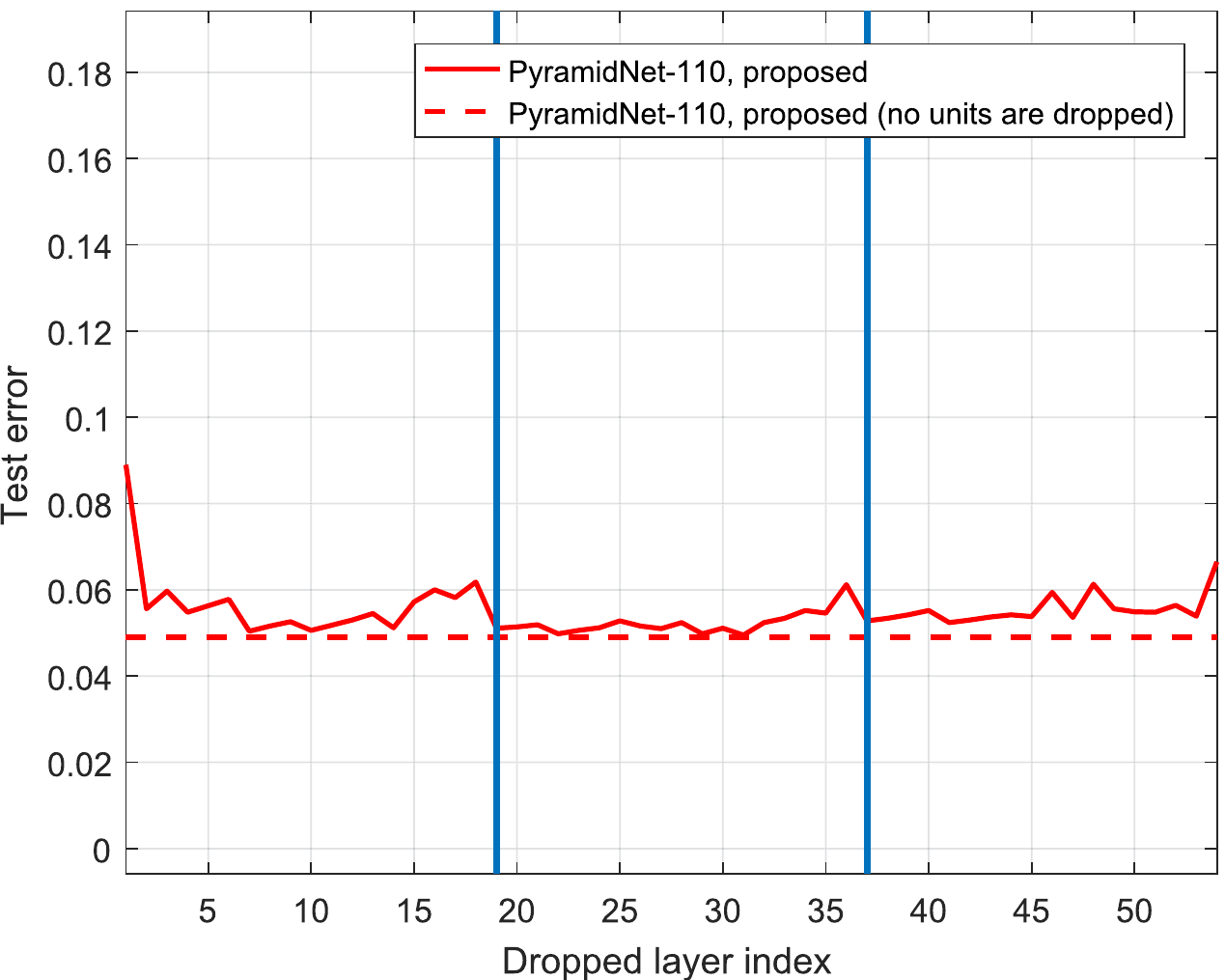}
\end{tabular}
\end{center}
\caption{Test error curves to study the extent to which residual units contribute to the performance in different network architectures by deleting their individual units. The dashed and solid lines denote the test errors that occur when no units are deleted, and when an individual unit is deleted, respectively. Bold vertical lines denote the location of residual units through downsampling.}
\label{fig:lines}
\end{figure}

Second, we verify the ensemble effect of our PyramidNets by evaluating the performance after deleting individual units, similar to the experiment of Veit~{\it et al.}~\cite{ensemble}. The results are shown in Figure~\ref{fig:lines}. As mentioned by Veit~{\it et al.}~\cite{ensemble}, removing individual units only causes a slight performance loss, compared with a plain network such as the VGG~\cite{VGG}. However, in the case of the pre-activation ResNet, removing the blocks subjected to downsampling tends to affect the classification accuracy by a relatively large margin, whereas this does not occur with our PyramidNets. Furthermore, the mean average error differences between the baseline result and the result obtained when individual units were deleted from both the pre-activation ResNet and our PyramidNet were 0.72\% and 0.54\%, respectively. This result shows that the ensemble effect of our PyramidNet becomes stronger than the original ResNet, such that generalization ability is improved.

\subsection{Zero-padded Shortcut Connection}
ResNets and pre-activation ResNets~\cite{resnet, preresnet} were studied several types of shortcuts, such as an identity-mapping shortcut or projection shortcut. The experimental results in~\cite{preresnet} showed that the identity-mapping shortcut is a much more appropriate choice than other shortcuts. Because an identity-mapping shortcut does not have parameters, it has a lower possibility of overfitting compared to the other types of shortcuts; this ensures improved generalization ability. Moreover, it can purely pass through the gradient according to the identity mapping, and therefore it provides more stability in the training stage.


In the case of our PyramidNet, identity mapping alone cannot be used for a shortcut because the feature map dimension differs among individual residual units. Therefore, only a zero-padded shortcut or projection shortcut can be used for all the residual units. However, as discussed in \cite{preresnet}, a projection shortcut can hamper information propagation and lead to optimization problems, especially for very deep networks. On the other hand, we found that the zero-padded shortcut does not lead to the overfitting problem because no additional parameters exist, and surprisingly, it shows significant generalization ability compared to other shortcuts.

\begin{figure}[t]
\begin{center}
\begin{tabular}{cc}
\includegraphics[trim = 4mm 0mm 4mm 0mm, clip, width=35mm, height = 31mm]{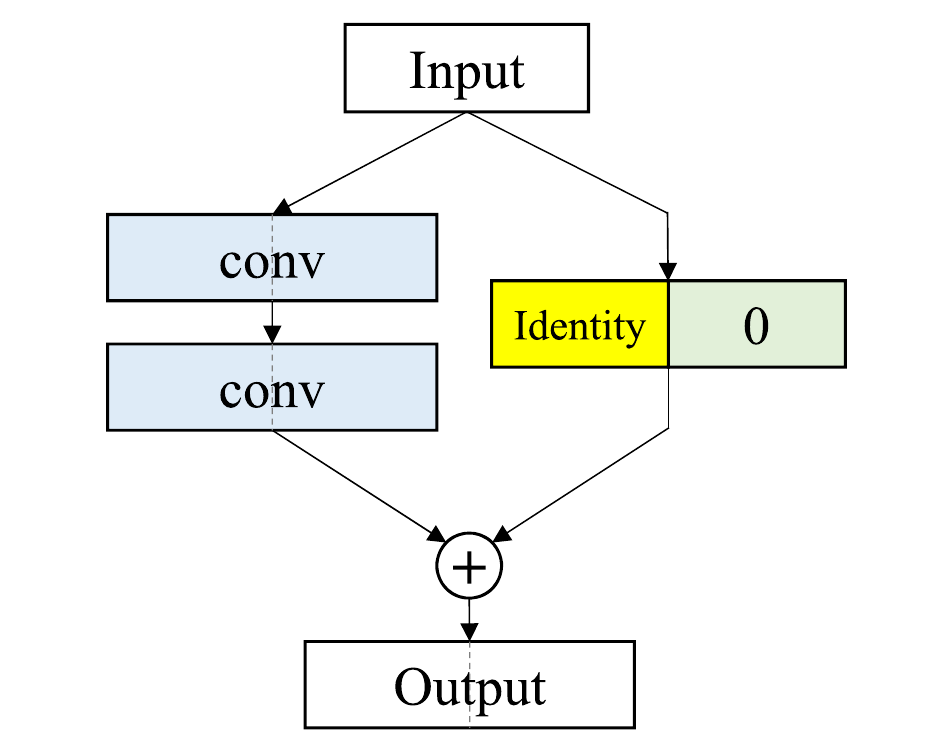} &
\includegraphics[trim = 0mm 0mm 0mm 0mm, clip,width=40mm, height = 31mm]{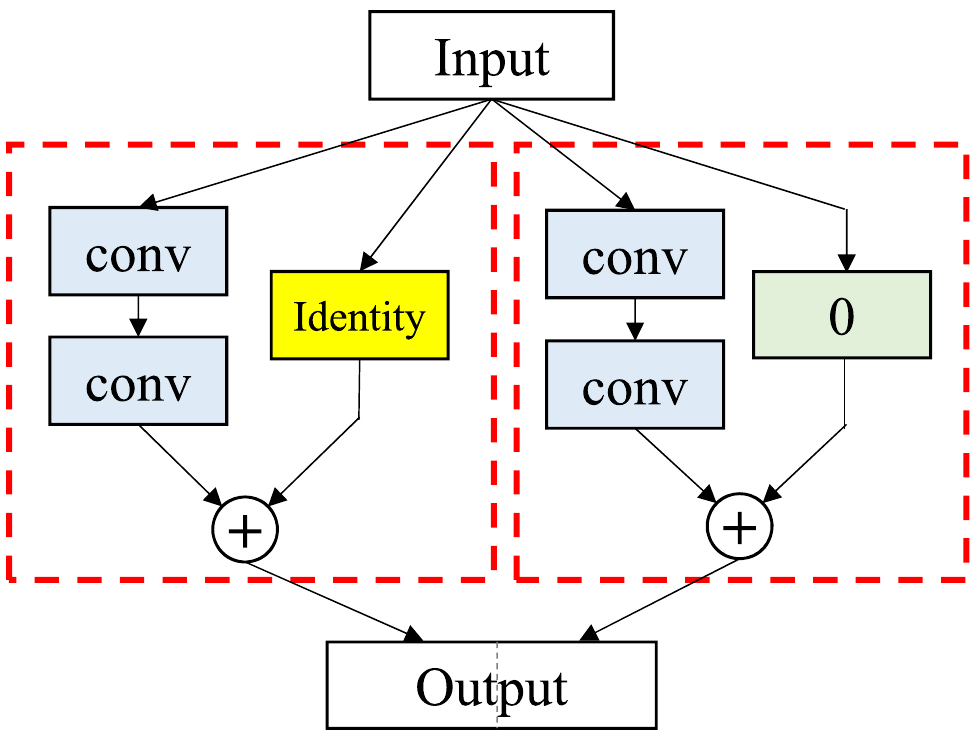} \\
(a) & \  (b)
\end{tabular}
\end{center}
\caption{Structure of residual unit (a) with zero-padded identity-mapping shortcut, (b) unraveled view of (a) showing that the zero-padded identity-mapping shortcut constitutes a mixture of a residual network with a shortcut connection and a plain network.}
\label{fig:zeropad}
\end{figure}

We now examine the effect of the zero-padded identity-mapping shortcut on the $k$-th residual unit that belongs to the $n$-th group with the reshaped vector $\mb{x}^{l}_{k}$ of the $l$-th feature map:
\bea
\mb{x}^{l}_{k}= \begin{cases}
                    \mb{F}_{(k,l)}(\mb{x}^{l}_{k-1}) + \mb{x}^{l}_{k-1}, &  \text{if $ 1 \leq l \leq D_{k-1}$}\\
                   \mb{F}_{(k,l)}(\mb{x}^{l}_{k-1}), &  \text{if $ D_{k-1} < l \leq D_{k} $}
                \end{cases}
\label{eq:zeropadded}
\eea
where $\mb{F}_{(k,l)}(\cdot)$ denotes the $l$-th residual function of the $k$-th residual unit and $D_{k}$ represents the pre-defined channel dimensions of the $k$-th residual unit. From eq.\eqref{eq:zeropadded}, zero-padded elements of the identity-mapping shortcut for increasing dimension let $\mb{x}^{l}_{k}$ contain the outputs of both residual networks and plain networks. Therefore, we could conjecture that each zero-padded identity-mapping shortcut can provide a mixture of the residual network and plain network, as shown in Figure~\ref{fig:zeropad}. Furthermore, our PyramidNet increases the channel dimension at every residual unit, and the mixture effect of the residual network and plain network increases markedly. Figure~\ref{fig:lines} supports the conclusion that the test error of PyramidNet does not oscillate as much as that of the pre-activation ResNet. Finally, we investigate several types of shortcuts including proposed zero-padded identity-mapping shortcut in Table~\ref{table:shortcut}.

\subsection{A New Building Block}
\label{ssec:newblock}
To maximize the capability of the network, it is natural to ask the following question: {\bf ``Can we design a better building block by altering the stacked elements inside the building block in more principled way?"}. The first building block types were proposed in the original paper on ResNets \cite{resnet}, and another type of building block was subsequently proposed in the paper on pre-activation ResNets \cite{preresnet}, to answer the question. Moreover, pre-activation ResNets attempted to solve the backward gradient flowing problem \cite{preresnet} by redesigning residual modules; this proved to be successful in trials. However, although the pre-activation residual unit was discovered with empirically improved performance, further investigation over the possible combinations is not yet performed, leaving a potential room for improvement. We next attempt to answer the question from two points of view by considering Rectified Linear Units (ReLUs)~\cite{ReLU} and Batch Normalization (BN)~\cite{BN} layers.


\begin{table}[t]
\footnotesize
\fontsize{7}{8}\selectfont
\begin{center}
\begin{tabular}{|l|c|c|}
\hline
Shortcut Types  & CIFAR-10 & CIFAR-100\\
\hline
(a) Identity mapping with projection shortcut & 5.03 & 23.48\\
(b) Projection with zero-padded shortcut & 6.84 & 31.29 \\
(c) Only projection shortcut & 6.98 & 31.62 \\
(d) Identity mapping with zero-padded shortcut & {\bf4.70}  & {\bf 22.77}\\
\hline
\end{tabular}
\end{center}
\caption{Top-1 errors (\%) on CIFAR datasets using our PyramidNet with several combinations of shortcut connections.}
\label{table:shortcut}
\end{table}

\begin{figure*}[t]
\begin{center}
\begin{tabular}{cccc}
\includegraphics[trim = 0mm 0mm 180mm 0mm, clip, width=0.21\textwidth, height=0.29\textwidth]{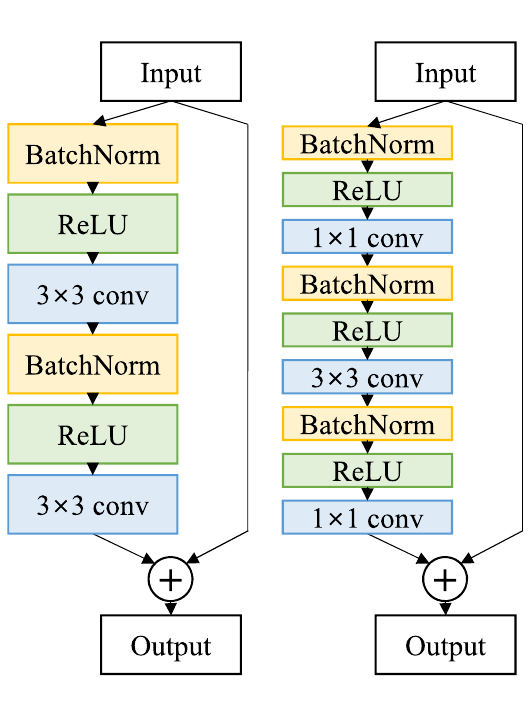} & \quad
\includegraphics[trim = 60mm 0mm 120mm 0mm, clip, width=0.21\textwidth, height=0.29\textwidth]{Images2/resnets2_final.pdf} & \quad
\includegraphics[trim = 120mm 0mm 60mm 0mm, clip, width=0.21\textwidth, height=0.29\textwidth]{Images2/resnets2_final.pdf} & \quad
\includegraphics[trim = 180mm 0mm 0mm 0mm, clip, width=0.21\textwidth, height=0.29\textwidth]{Images2/resnets2_final.pdf} \\
\quad  (a) & \quad \quad (b) & \quad \quad  (c) & \quad \quad (d)
\end{tabular}
\end{center}
\caption{Various types of basic and bottleneck residual units. ``BatchNorm" denotes a Batch Normalization (BN) layer. (a) original pre-activation ResNets~\cite{preresnet}, (b) pre-activation ResNets removing the first ReLU, (c) pre-activation ResNets with a BN layer after the final convolutional layer, and (d) pre-activation ResNets removing the fist ReLU with a BN layer after the final convolutional layer.}
\label{fig:resnets}
\end{figure*}

\subsubsection{ReLUs in a Building Block}
Including ReLUs~\cite{ReLU} in the building blocks of residual units is essential for nonlinearity; however, we found empirically that the performance can vary depending on the locations and the number of ReLUs. This could be discussed with original ResNets \cite{resnet}, for which it was shown that the performance increases as the network becomes deeper; however, if the depth exceeds 1,000 layers, overfitting still occurs and the result is less accurate than that generated by shallower ResNets.

First, we note that using ReLUs after the addition of residual units adversely affects performance:
\bea
\mb{x}^{l}_{k} = \it{ReLU}(\mb{F}_{(k,l)}(\mb{x}^{l}_{k-1}) + \mb{x}^{l}_{k-1}),
\eea
where the ReLUs seem to have the function of filtering non-negative elements. Gross~and~Wilber~\cite{torchblog} found that simply removing ReLUs from the original ResNet~\cite{resnet} after each addition with the shortcut connection leads to small performance improvements. This could be understood by considering that, after addition, ReLUs provide non-negative input to the subsequent residual units, and therefore the shortcut connection is always non-negative and the convolutional layers would take responsibility for producing negative output before addition; this may decrease the overall capability of the network architecture as analyzed in \cite{preresnet}. The pre-activation ResNets proposed by He~{\it et al.}~\cite{preresnet} also overcame this issue with pre-activated residual units that place BN layers and ReLUs before (instead of after) the convolutional layers:
\bea
\mb{x}^{l}_{k} = \mb{F}_{(k,l)}(\mb{x}^{l}_{k-1}) + \mb{x}^{l}_{k-1},
\eea
where ReLUs are removed after addition to create an identity path. Consequently, the overall performance has increased by a large margin without overfitting, even at depths exceeding 1,000 layers. Furthermore, Shen~{\it et al.}~\cite{weightedresnet} proposed a weighted residual network architecture, which locates a ReLU inside a residual unit (instead of locating ReLU after addition) to create an identity path, and showed that this structure also does not overfit even at depths of more than 1,000 layers.

Second, we found that the use of a large number of ReLUs in the blocks of each residual unit may negatively affect performance. Removing the first ReLU in the blocks of each residual unit, as shown in Figure~\ref{fig:resnets}~(b) and (d), was found to enhance performance compared with the blocks shown in Figure~\ref{fig:resnets}~(a) and (c). Experimentally, we found that removal of the first ReLU in the stack is preferable and that the other ReLU should remain to ensure nonlinearity.
Removing the second ReLU in Figure~\ref{fig:resnets}~(a) changes the blocks to {\it BN-ReLU-conv-BN-conv}, and it is clear that, in these blocks, the convolutional layers are successively located without ReLUs to weaken their representation powers of each other. However, when we remove the first ReLU, the blocks are changed to {\it BN-conv-BN-ReLU-conv}, in which case the two convolutional layers are separated by the second ReLU, thereby guaranteeing nonlinearity.
The results in Table \ref{table:compare} confirm that removing the first ReLU as in (b) and (d) in Figure \ref{fig:resnets}, enhances the performance. Consequently, provided that an appropriate number of ReLUs are used to guarantee the nonlinearity of the feature space manifold, the remaining ReLUs could be removed to improve network performance.

\subsubsection{BN Layers in a Building Block}
The main role of a BN layer is to normalize the activations for fast convergence and to improve performance.
The experimental results of the four structures provided in Table~\ref{table:compare} show that the BN layer can be used to maximize the capability of a single residual unit. A BN layer conducts an affine transformation with the following equation:

\begin{table}[t]
\fontsize{7}{8}\selectfont
\begin{center}
\begin{tabular}{|l|c|c|}
\hline
ResNet Architecture  & CIFAR-10 & CIFAR-100\\
\hline
(a) Pre-activation~\cite{preresnet} & 5.82 & 25.06\\
(b) Removing the first ReLU & 5.31 & 24.55\\
(c) BN after the final conv & 5.74 & 24.54 \\
(d) (b) + (c) & 5.29 & 23.74 \\
\hline\hline
PyramidNet Architecture & CIFAR-10 & CIFAR-100\\
\hline
(a) Pre-activation~\cite{preresnet} & 5.15 & 24.40\\
(b) Removing the first ReLU & 4.81 & 23.43\\
(c) BN after the final conv & 4.96 & 23.89\\
(d) (b) + (c) & {\bf4.62} & {\bf23.31}\\
\hline\hline
PyramidNet (bottleneck) Architecture  & CIFAR-10 & CIFAR-100\\
\hline
(a) Pre-activation~\cite{preresnet} & 4.61 & 21.10\\
(b) Removing the first ReLU  & 4.45 & 20.40\\
(c) BN after the final conv & 4.56 & 20.44\\
(d) (b) + (c)  & {\bf4.26} & {\bf20.32}\\
\hline
\end{tabular}
\end{center}
\caption{Top-1 errors (\%) on CIFAR datasets for several building block combinations of ReLUs and BN layers shown in Figure~\ref{fig:resnets}~(a)--(d), using ResNet~\cite{preresnet} (with original feature map dimension configuration) and our PyramidNet.}
\label{table:compare}
\vspace{-3mm}
\end{table}
\begin{table*}[t]
\fontsize{9}{10}\selectfont
\begin{center}
\begin{tabular}{|l|c|c|c|c|c|c|}
\hline
Network & \# of Params & Output Feat. Dim. &  Depth & Training Mem. & CIFAR-10 & CIFAR-100 \\
\hline\hline
NiN~\cite{NiN} & - & - & - & - & 8.81& 35.68\\
All-CNN~\cite{allcnn} & - & - & - & - & 7.25& 33.71\\
DSN~\cite{DSN} & - & - & - & - & 7.97 & 34.57 \\
FitNet~\cite{fitnet} & - & - & - & - & 8.39 & 35.04 \\
Highway~\cite{Highway} & - & - & - & - & 7.72 & 32.39 \\
Fractional Max-pooling~\cite{fracc} & - & - & -& - & 4.50 & 27.62 \\
ELU~\cite{Highway} & - & - & - & - & 6.55 & 24.28 \\
\hline
ResNet~\cite{resnet} & 1.7M & 64 & 110 & 547MB & 6.43& 25.16\\
ResNet~\cite{resnet} & 10.2M & 64 & 1001 & 2,921MB & - & 27.82\\
ResNet~\cite{resnet} & 19.4M & 64 & 1202 & 2,069MB & 7.93& -\\
\hline
Pre-activation ResNet~\cite{preresnet} & 1.7M & 64 & 164 & 841MB & 5.46& 24.33\\
Pre-activation ResNet~\cite{preresnet} & 10.2M & 64 & 1001 & 2,921MB & 4.62& 22.71\\
\hline
Stochastic Depth~\cite{stochasticdepth} & 1.7M & 64 &  110 & 547MB & 5.23& 24.58\\
Stochastic Depth~\cite{stochasticdepth} & 10.2M & 64 & 1202 & 2,069MB & 4.91& -\\
\hline
FractalNet~\cite{fractalnet} & 38.6M & 1,024 & 21 & - & 4.60& 23.73\\
\hline
SwapOut v2 (width$\times4$)~\cite{swapout} & 7.4M & 256 & 32 & - & 4.76& 22.72\\
\hline
Wide ResNet (width$\times4$)~\cite{wideresnet} & 8.7M & 256 & 40 & 775MB & 4.97& 22.89\\
Wide ResNet (width$\times10$)~\cite{wideresnet} & 36.5M & 640 & 28 & 1,383MB & 4.17& 20.50\\
\hline
Weighted ResNet~\cite{weightedresnet} & 19.1M & 64 & 1192 & - & 5.10& -\\
\hline
DenseNet ($k=24$)~\cite{densenet} & 27.2M & 2,352 & 100 & 4,381MB & 3.74 & 19.25 \\
DenseNet-BC ($k=40$)~\cite{densenet} & 25.6M & 2,190 & 190 & 7,247MB & 3.46 & 17.18 \\
\hline
PyramidNet ($\alpha=48$)             & 1.7M  & 64  & 110 & 655MB & 4.58$\pm$0.06& 23.12$\pm$0.04\\
PyramidNet ($\alpha=84$)             & 3.8M  & 100 & 110 & 781MB & 4.26$\pm$0.23 & 20.66$\pm$0.40\\
PyramidNet ($\alpha=270$)            & 28.3M & 286 & 110 & 1,437MB & {3.73}$\pm$0.04 & {18.25}$\pm$0.10\\
PyramidNet (bottleneck, $\alpha=270$)& 27.0M & 1,144 & 164 & 4,169MB & {3.48}$\pm$0.20 & {17.01}$\pm$0.39\\
PyramidNet (bottleneck, $\alpha=240$)& 26.6M & 1,024 & 200 & 4,451MB & {3.44}$\pm$0.11 & {16.51}$\pm$0.13\\
PyramidNet (bottleneck, $\alpha=220$)& 26.8M & 944 & 236 & 4,767MB & {3.40}$\pm$0.07 & {16.37}$\pm$0.29\\
PyramidNet (bottleneck, $\alpha=200$)& 26.0M & 864 & 272 & 5,005MB & {\bf \color{red} 3.31}$\pm$0.08 & {\bf \color{red} 16.35}$\pm$0.24\\
\hline
\end{tabular}
\end{center}
\caption{Top-1 error rates (\%) on CIFAR datasets. All the results of PyramidNets are produced with additive PyramidNets, and $\alpha$ denotes the widening factor. ``Output Feat. Dim.'' denotes the feature dimension of just before the last softmax classifier. The best results are highlighted in {\bf \color{red} red}.}
\label{table:result}
\end{table*}
\begin{equation}
\mb{y} =\gamma \mb{x} + \beta,
\end{equation}
where $\gamma$ and $\beta$ are learned for every activation in feature maps. We experimentally found that the learned $\gamma$ and $\beta$ could closely approximate $0$. This implies that if the learned $\gamma$ and $\beta$ are both close to $0$, then the corresponding activation is considered not to be useful. Weighted ResNets \cite{weightedresnet}, in which the learnable weights occur at the end of their building blocks, are also similarly learned to determine whether the corresponding residual unit is useful. Thus, the BN layers at the end of each residual unit are a generalized version including \cite{weightedresnet} to enable decisions to be made as to whether each residual unit is helpful. Therefore, the degrees of freedom obtained by involving $\gamma$ and $\beta$ from the BN layers could improve the capability of the network architecture. The results in Table \ref{table:compare} support the conclusion that adding a BN layer at the end of each building block, as in type (c) and (d) in Figure \ref{fig:resnets}, improves the performance. Note that the aforementioned network removing the first ReLU is also improved by adding a BN layer after the final convolutional layer. Furthermore, the results in Table \ref{table:compare} show that both PyramidNet and a new building block improve the performance significantly.


\section{Experimental Results}
\label{section:exp}




We evaluate and compare the performance of our algorithm with that of existing algorithms~\cite{resnet,preresnet,NiN,weightedresnet,wideresnet} using representative benchmark datasets: CIFAR-10 and CIFAR-100~\cite{cifar}. CIFAR-10 and CIFAR-100 each contain 32$\times$32-pixel color images, consists of 50,000 training images and 10,000 testing images. But in case of CIFAR-10, it includes 10 classes, and CIFAR-100 includes 100 classes. The standard data augmentation, horizontal flipping, and translation by 4 pixels are adopted in our experiments, following the common practice~\cite{NiN}. The results achieved by PyramidNets are based on the proposed residual unit: placing a BN layer after the final convolutional layer, and removing the first ReLU as in Figure~\ref{fig:resnets}~(d). Our code is built on Torch open source deep learning framework~\cite{torch7}.

\subsection{Training Settings}

Our PyramidNets are trained using backpropagation~\cite{backprop} by Stochastic Gradient Descent (SGD) with Nesterov momentum for 300 epochs on CIFAR-10 and CIFAR-100 datasets. The initial learning rate is set to 0.1 for CIFAR-10 and 0.5 for CIFAR-100, and is decayed by a factor of 0.1 at 150 and 225 epochs, respectively. The filter parameters are initialized by ``msra"~\cite{prelu}. We use a weight decay of 0.0001, a dampening of 0, a momentum of 0.9, and a batch size of 128.
\begin{figure}[t]
\begin{center}
\begin{tabular}{cc}
\includegraphics[trim = 40mm 85mm 45mm 94mm, clip,  width=0.2\textwidth, height=0.175\textwidth]{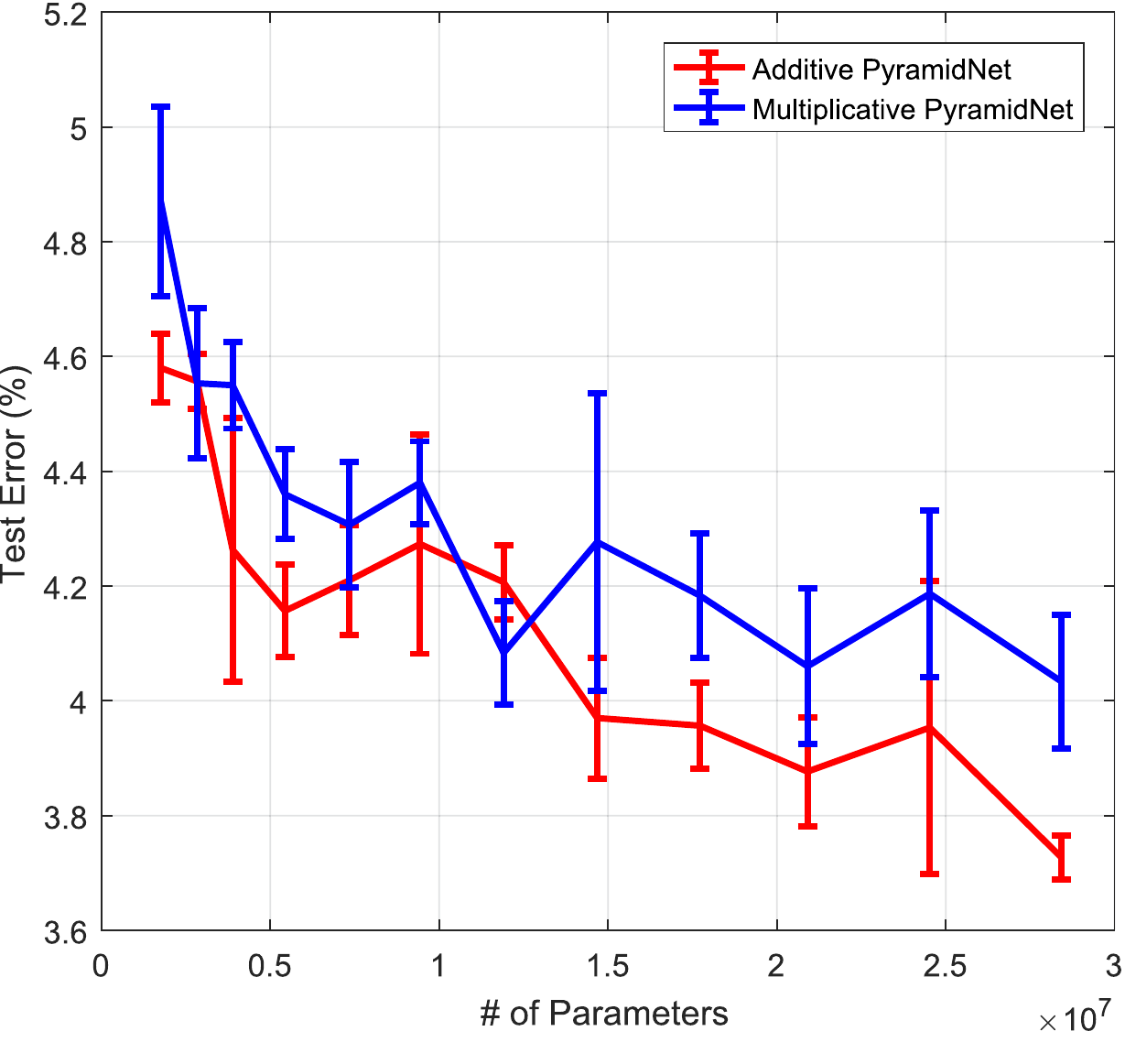} \ \quad
\includegraphics[trim = 40mm 85mm 45mm 94mm, clip,  width=0.2\textwidth, height=0.175\textwidth]{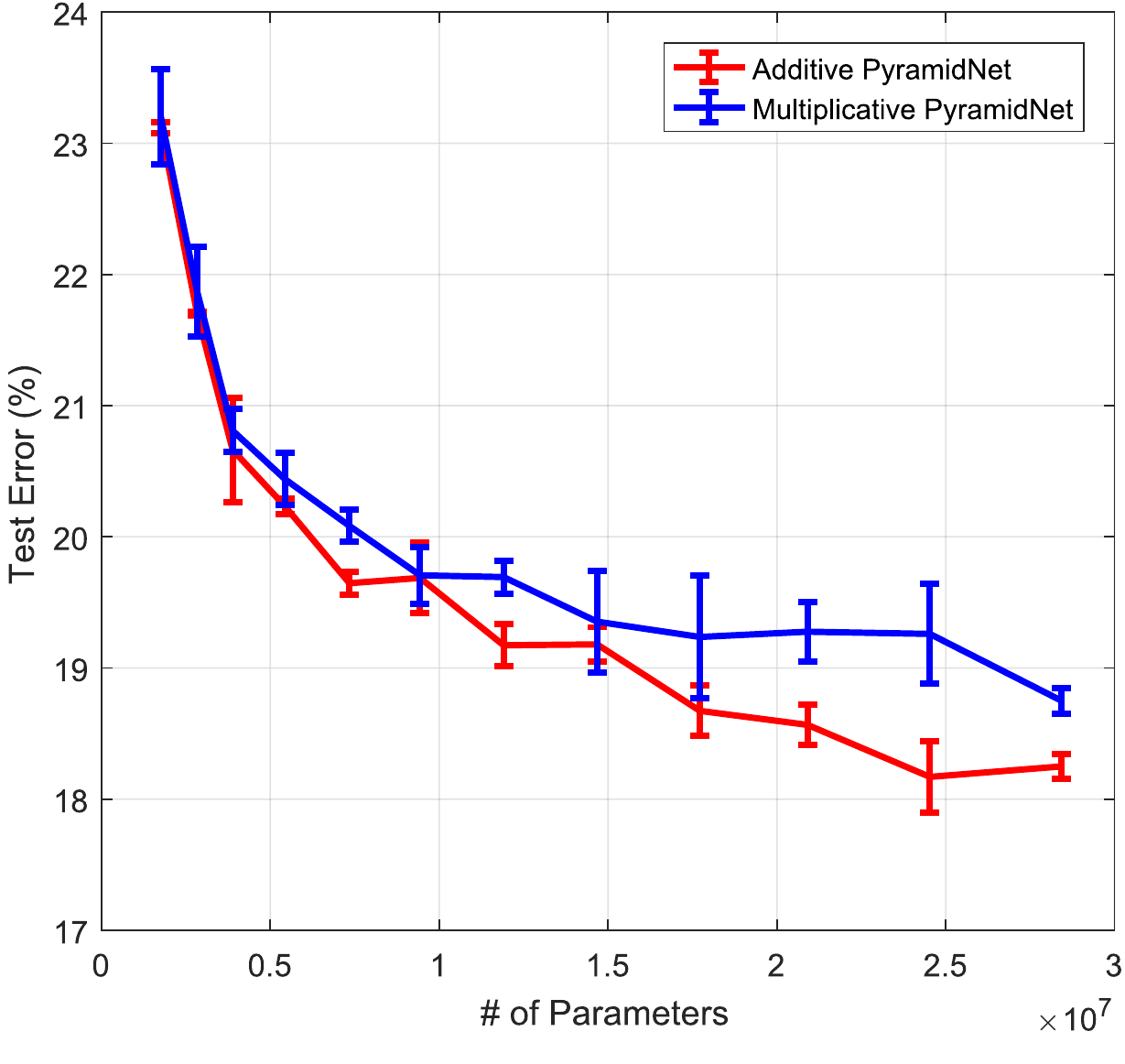}
\end{tabular}
\end{center}
\caption{Comparison of test error curves with error bars of additive PyramidNet and multiplicative PyramidNet on CIFAR-10 (left) and CIFAR-100 (right) datasets, according to the different number of parameters.}
\label{fig:addmulcomparision}
\vspace{-3mm}
\end{figure}

\begin{table*}[t]
\fontsize{9}{10}\selectfont
\begin{center}
\centering
\begin{tabular}{|l|c|c|c|c|c|c|c|}
\hline
Network & \# of Params & Output Feat. Dim. & Augmentation & Train Crop & Test Crop & Top-1 & Top-5 \\
\hline\hline
ResNet-152~\cite{resnet} & 60.0M & 2,048 & scale & 224$\times$224 & 224$\times$224 & 23.0 & 6.7 \\
Pre-ResNet-152$^\dagger$~\cite{preresnet} & 60.0M & 2,048 & scale+asp ratio & 224$\times$224 & 224$\times$224 & 22.2 & 6.2\\
Pre-ResNet-200$^\dagger$~\cite{preresnet} & 64.5M & 2,048 & scale+asp ratio & 224$\times$224 & 224$\times$224 & 21.7 & 5.8\\
WRN-50-2-bottleneck~\cite{wideresnet} & 68.9M & 2,048 & scale+asp ratio & 224$\times$224 & 224$\times$224 & 21.9 & 6.0\\
PyramidNet-200 ($\alpha=300$) & 62.1M & 1,456 & scale+asp ratio & 224$\times$224 & 224$\times$224 & {\bf20.5} & {\bf5.3}\\
PyramidNet-200 ($\alpha=300$)$^*$ & 62.1M & 1,456 & scale+asp ratio & 224$\times$224 & 224$\times$224 & {\bf20.5} & {\bf5.4}\\
PyramidNet-200 ($\alpha=450$)$^*$ & 116.4M & 2,056 & scale+asp ratio & 224$\times$224 & 224$\times$224 & {\bf20.1} & {\bf5.4}\\
\hline
ResNet-200~\cite{resnet} & 64.5M & 2,048 & scale & 224$\times$224 & 320$\times$320 & 21.8 & 6.0\\
Pre-ResNet-200~\cite{preresnet} & 64.5M & 2,048 & scale+asp ratio & 224$\times$224 & 320$\times$320 & 20.1 & 4.8\\
Inception-v3~\cite{Inceptionv3} & - & 2,048 & scale+asp ratio  & 299$\times$299 & 299$\times$299 & 21.2 & 5.6\\
Inception-ResNet-v1~\cite{InceptionResnet}  &  - & 1,792 & scale+asp ratio & 299$\times$299 & 299$\times$299 & 21.3 & 5.5\\
Inception-v4~\cite{InceptionResnet}  & - & 1,536 & scale+asp ratio & 299$\times$299 & 299$\times$299 & 20.0 & 5.0\\
Inception-ResNet-v2~\cite{InceptionResnet}  &  - & 1,792 & scale+asp ratio & 299$\times$299 & 299$\times$299 & 19.9 & 4.9\\
PyramidNet-200 ($\alpha=300$) & 62.1M & 1,456 & scale+asp ratio & 224$\times$224 & 320$\times$320 & {\bf19.6} & {\bf4.8}\\
PyramidNet-200 ($\alpha=300$)$^*$ & 62.1M & 1,456 & scale+asp ratio & 224$\times$224 & 320$\times$320 & {\bf19.5} & {\bf4.8}\\
PyramidNet-200 ($\alpha=450$)$^*$ & 116.4M & 2,056 & scale+asp ratio & 224$\times$224 & 320$\times$320 & {\bf19.2} & {\bf4.7}\\
\hline
\end{tabular}
\end{center}
\caption{Comparisons of single-model, single-crop error (\%) on the ILSVRC 2012 validation set. All the results of PyramidNets are produced with additive PyramidNets. ``asp ratio" means the aspect ratio applied for data augmention, and ``Output feat. dim." denotes the feature dimension of just after the last global pooling layer. $^*$ denotes the models which applied dropout method, and $\dagger$ denotes the results obtained from {\it https://github.com/facebook/fb.resnet.torch}.}
\label{table:imagenetres}
\end{table*}

\subsection{Performance Evaluation}

In our work, we mainly use the top-1 error rate for evaluating our network architecture. Additive PyramidNets with both basic and pyramidal bottleneck residual units are used. The error rates are provided in Table~\ref{table:result} for ours and the state-of-the-art models. The experimental results show that our network has superior generalization ability, in terms of the number of parameters, showing the best results compared with other models. 

Figure~\ref{fig:addmulcomparision} compares additive and multiplicative PyramidNets using CIFAR datasets. When the number of parameters is low, both additive and multiplicative PyramidNets show similar performance, because these two network architectures do not have significant structural differences. As the number of parameters increases, they start to show a more marked difference in terms of the feature map dimension configuration. Because the feature map dimension increases linearly in the case of additive PyramidNets, the feature map dimensions of the input-side layers tend to be larger, and those of the output-side layers tend to be smaller, compared with multiplicative PyramidNets as illustrated in Figure~\ref{fig:addmul}.

Previous works~\cite{resnet, VGG} typically set multiplicative scaling of feature map dimension for downsampling modules, which is implemented to give a larger degree of freedom to the classification part by increasing the feature map dimension of the output-side layers. However, for our PyramidNet, the results in Figure~\ref{fig:addmulcomparision} implies that increasing the model capacity of the input-side layers would lead to a better performance improvement than using a conventional way of multiplicative scaling of feature map dimension.

We also note that, although the use of regularization methods such as dropout~\cite{dropout} or stochastic depth~\cite{stochasticdepth} could further improve the performance of our model, we did not involve those methods to ensure a fair comparison with other models.

\subsection{ImageNet}
1,000-class ImageNet dataset~\cite{ImageNet} used for ILSVRC contains more than one million training images and 50,000 validation images. We use additive PyramidNets with the pyramidal bottleneck residual units, deleting the first ReLU and adding a BN layer at the last layer as described in Section~\ref{ssec:newblock} and shown in Figure~\ref{fig:resnets} (d) for further performance improvement.

We train our models for 120 epochs with a batch size of 128, and the initial learning rate is set to 0.05, divided by 10 at 60, 90 and 105 epochs. We use the same weight decay, momentum, and initialization settings as those of CIFAR datasets. We train our model by using a standard data augmentation with scale jittering and aspect ratio as suggested in Szegedy~{\it et al.}~\cite{GoogleNet}. Table~\ref{table:imagenetres} shows the results of our PyramidNets in ImageNet dataset compared with the state-of-the-art models. 
The experimental results show that our PyramidNet with $\alpha=300$ has a top-1 error rate of 20.5$\%$, which is 1.2$\%$ lower than the pre-activation ResNet-200~\cite{preresnet} which has a similar number of parameters but higher output feature dimension than our model. We also notice that increasing $\alpha$ with an appropriate regularization method can further improve the performance.

For comparison with the Inception-ResNet~\cite{InceptionResnet} that uses a testing crop with $299\times 299$ size, we test our model on a $320\times 320$ crop, by the same reason with the work of He~{\it et al.}~\cite{preresnet}. Our PyramidNet with $\alpha=300$ shows a top-1 error rate of 19.6$\%$, which outperforms both the pre-activation ResNet~\cite{preresnet} and the Inception-ResNet-v2~\cite{InceptionResnet} models.
\section{Conclusion}
The main idea of the novel deep network architecture described in this paper involves increasing the feature map dimension gradually, in order to construct so-called PyramidNets along with the concept of ResNets. We also developed a novel residual unit, which includes a new building block for a residual unit with a zero-padded shortcut; this design leads to significantly improved generalization ability. In tests using CIFAR-10, CIFAR-100, and ImageNet-1k datasets, our PyramidNets outperform all previous state-of-the-art deep network architectures. Furthermore, the insights in this paper could be utilized by any network architecture, to improve their capacity for better performance. In future work, we will develop methods of optimizing parameters such as feature map dimensions in more principled ways with proper cost functions that give insight into the nature of residual networks.
\label{sec:conclusion}

{\noindent  \bf Acknowledgements:}
{
This work was supported by the ICT R\&D program of MSIP/IITP, 2016-0-00563, Research on Adaptive Machine Learning Technology Development for Intelligent Autonomous Digital Companion.
}

{\small
\bibliographystyle{ieee}
\bibliography{egbib}

\begin{thebibliography}{10}\itemsep=-1pt

\bibitem{torch7}
R.~Collobert, K.~Kavukcuoglu, and C.~Farabet.
\newblock Torch7: A matlab-like environment for machine learning.
\newblock In {\em BigLearn, NIPS Workshop}, 2011.

\bibitem{decaf}
J.~Donahue, Y.~Jia, O.~Vinyals, J.~Hoffman, N.~Zhang, E.~Tzeng, and T.~Darrell.
\newblock Decaf: A deep convolutional activation feature for generic visual
  recognition.
\newblock In {\em ICML}, 2014.

\bibitem{rcnn}
R.~Girshick, J.~Donahue, T.~Darrell, and J.~Malik.
\newblock Rich feature hierarchies for accurate object detection and semantic
  segmentation.
\newblock In {\em CVPR}, 2014.

\bibitem{fracc}
B.~Graham.
\newblock Fractional max-pooling.
\newblock {\em arXiv preprint arXiv:1412.6071}, 2014.

\bibitem{torchblog}
S.~Gross and M.~Wilber.
\newblock Training and investigating residual nets.
\newblock 2016.
\newblock http://torch.ch/blog/2016/02/04/resnets.html.

\bibitem{prelu}
K.~He, X.~Zhang, S.~Ren, and J.~Sun.
\newblock Delving deep into rectifiers: Surpassing human-level performance on
  imagenet classification.
\newblock In {\em ICCV}, 2015.

\bibitem{resnet}
K.~He, X.~Zhang, S.~Ren, and J.~Sun.
\newblock Deep residual learning for image recognition.
\newblock In {\em CVPR}, 2016.

\bibitem{preresnet}
K.~He, X.~Zhang, S.~Ren, and J.~Sun.
\newblock Identity mappings in deep residual networks.
\newblock In {\em ECCV}, 2016.

\bibitem{densenet}
G.~Huang, Z.~Liu, and K.~Q. Weinberger.
\newblock Densely connected convolutional networks.
\newblock {\em arXiv preprint arXiv:1608.06993}, 2016.

\bibitem{stochasticdepth}
G.~Huang, Y.~Sun, Z.~Liu, D.~Sedra, and K.~Weinberger.
\newblock Deep networks with stochastic depth.
\newblock In {\em ECCV}, 2016.

\bibitem{BN}
S.~Ioffe and C.~Szegedy.
\newblock Batch normalization: Accelerating deep network training by reducing
  internal covariate shift.
\newblock In {\em ICML}, 2015.

\bibitem{cifar}
A.~Krizhevsky.
\newblock Learning multiple layers of features from tiny images.
\newblock In {\em Tech Report}, 2009.

\bibitem{alexnet}
A.~Krizhevsky, I.~Sutskever, and G.~E. Hinton.
\newblock {ImageNet Classification with Deep Convolutional Neural Networks}.
\newblock In {\em NIPS}, 2012.

\bibitem{fractalnet}
G.~Larsson, M.~Maire, and G.~Shakhnarovich.
\newblock Fractalnet: Ultra-deep neural networks without residuals.
\newblock {\em arXiv preprint arXiv:1605.07648}, 2016.

\bibitem{backprop}
Y.~LeCun, B.~Boser, J.~S. Denker, D.~Henderson, R.~E. Howard, W.~Hubbard, and
  L.~D. Jackel.
\newblock Backpropagation applied to handwritten zip code recognition.
\newblock {\em Neural computation}, 1(4):541--551, 1989.

\bibitem{Lenet}
Y.~LeCun, L.~Bottou, Y.~Bengio, and P.~Haffner.
\newblock Gradient-based learning applied to document recognition.
\newblock {\em Proceedings of the IEEE}, 86(11):2278--2324, 1998.

\bibitem{DSN}
C.-Y. Lee, S.~Xie, P.~Gallagher, Z.~Zhang, and Z.~Tu.
\newblock Deeply-supervised nets.
\newblock In {\em AISTATS}, 2015.

\bibitem{NiN}
M.~Lin, Q.~Chen, and S.~Yan.
\newblock Network in network.
\newblock In {\em ICLR}, 2014.

\bibitem{FCN}
J.~Long, E.~Shelhamer, and T.~Darrell.
\newblock Fully convolutional networks for semantic segmentation.
\newblock In {\em CVPR}, 2015.

\bibitem{ReLU}
V.~Nair and G.~E. Hinton.
\newblock Rectified linear units improve restricted boltzmann machines.
\newblock In {\em ICML}, 2010.

\bibitem{fitnet}
A.~Romero, N.~Ballas, S.~E. Kahou, A.~Chassang, C.~Gatta, and Y.~Bengio.
\newblock Fitnets: Hints for thin deep nets.
\newblock In {\em ICLR}, 2015.

\bibitem{ImageNet}
O.~Russakovsky, J.~Deng, H.~Su, J.~Krause, S.~Satheesh, S.~Ma, Z.~Huang,
  A.~Karpathy, A.~Khosla, M.~Bernstein, A.~C. Berg, and L.~Fei-Fei.
\newblock {ImageNet Large Scale Visual Recognition Challenge}.
\newblock {\em International Journal of Computer Vision}, 115(3):211--252,
  2015.

\bibitem{overfeat}
P.~Sermanet, D.~Eigen, X.~Zhang, M.~Mathieu, R.~Fergus, and Y.~LeCun.
\newblock Overfeat: Integrated recognition, localization and detection using
  convolutional networks.
\newblock In {\em ICLR}, 2014.

\bibitem{weightedresnet}
F.~Shen and G.~Zeng.
\newblock Weighted residuals for very deep networks.
\newblock {\em arXiv preprint arXiv:1605.08831}, 2016.

\bibitem{VGG}
K.~Simonyan and A.~Zisserman.
\newblock Very deep convolutional networks for large-scale image recognition.
\newblock In {\em ICLR}, 2015.

\bibitem{swapout}
S.~Singh, D.~Hoiem, and D.~Forsyth.
\newblock Swapout: Learning an ensemble of deep architectures.
\newblock In {\em NIPS}, 2016.

\bibitem{allcnn}
J.~T. Springenberg, A.~Dosovitskiy, T.~Brox, and M.~Riedmiller.
\newblock Striving for simplicity: The all convolutional net.
\newblock In {\em ICLR Workshop}, 2015.

\bibitem{dropout}
N.~Srivastava, G.~Hinton, A.~Krizhevsky, I.~Sutskever, and R.~Salakhutdinov.
\newblock Dropout: A simple way to prevent neural networks from overfitting.
\newblock {\em Journal of Machine Learning Research}, 15:1929--1958, 2014.

\bibitem{Highway}
R.~K. Srivastava, K.~Greff, and J.~Schmidhuber.
\newblock Training very deep networks.
\newblock In {\em NIPS}, 2015.

\bibitem{InceptionResnet}
C.~Szegedy, S.~Ioffe, and V.~Vanhoucke.
\newblock Inception-v4, inception-resnet and the impact of residual connections
  on learning.
\newblock In {\em ICLR Workshop}, 2016.

\bibitem{GoogleNet}
C.~Szegedy, W.~Liu, Y.~Jia, P.~Sermanet, S.~Reed, D.~Anguelov, D.~Erhan,
  V.~Vanhoucke, and A.~Rabinovich.
\newblock Going deeper with convolutions.
\newblock In {\em CVPR}, 2015.

\bibitem{Inceptionv3}
C.~Szegedy, V.~Vanhoucke, S.~Ioffe, J.~Shlens, and Z.~Wojna.
\newblock Rethinking the inception architecture for computer vision.
\newblock In {\em CVPR}, 2016.

\bibitem{ensemble}
A.~Veit, M.~Wilber, and S.~Belongie.
\newblock Residual networks behave like ensembles of relatively shallow
  networks.
\newblock In {\em NIPS}, 2016.

\bibitem{wideresnet}
S.~Zagoruyko and N.~Komodakis.
\newblock Wide residual networks.
\newblock In {\em BMVC}, 2016.

\bibitem{zfnet}
M.~D. Zeiler and R.~Fergus.
\newblock Visualizing and understanding convolutional networks.
\newblock In {\em ECCV}, 2014.

\end{thebibliography}
}

\end{document}